\documentclass[journal]{IEEEtran}
\IEEEoverridecommandlockouts
\usepackage{amsmath,amssymb,amsfonts}
\usepackage{algorithmic}
\usepackage{graphicx, hyperref, url}
\usepackage{textcomp}
\usepackage{csquotes}
\usepackage{xcolor,soul}
\usepackage{adjustbox}
\usepackage{hyperref}
\usepackage{multirow}
\usepackage{titlesec}
\usepackage{subcaption}
\usepackage{makecell}
\usepackage{amsfonts}

\usepackage[numbers,sort&compress]{natbib}

\usepackage{etoolbox}

\definecolor{columbiablue}{rgb}{0.61, 0.87, 1.0}
\titlespacing{\subsubsection}{0pt}{0pt}{1pt}

\usepackage{amsmath}

\usepackage{mathtools}
\usepackage[colorinlistoftodos,prependcaption,textsize=tiny]{todonotes}
\usepackage{gensymb}
\usepackage{multirow}
\usepackage{orcidlink}
\usepackage{parskip}

\begin{document}
\title{Adaptive Neural Unscented Kalman Filter}
\author{Amit Levy \orcidlink{0009-0009-8703-5807},
Itzik~Klein \orcidlink{0000-0001-7846-0654}
\thanks{A. Levy and I. Klein are with the Hatter Department of Marine Technologies, Charney School of Marine Sciences, University of Haifa, Israel.\\ Corresponding author: alevy02@campus.haifa.ac.il}
}

\maketitle
\begin{abstract}
    The unscented Kalman filter is an algorithm capable of handling nonlinear scenarios. Uncertainty in process noise covariance may decrease the filter estimation performance or even lead to its divergence. Therefore, it is important to adjust the process noise covariance matrix in real time. In this paper, we developed an adaptive neural unscented Kalman filter to cope with time-varying uncertainties during platform operation. To this end, we devised ProcessNet, a simple yet efficient end-to-end regression network to adaptively estimate the process noise covariance matrix. We focused on the nonlinear inertial sensor and Doppler velocity log fusion problem in the case of autonomous underwater vehicle navigation. Using a real-world recorded dataset from an autonomous underwater vehicle, we demonstrated our filter performance and showed its advantages over other adaptive and non-adaptive nonlinear filters.
\end{abstract}

\section{Introduction}\label{intro_sec}
The linear Kalman filter (KF) is an optimal minimum mean squared error (MMSE) estimator for state estimation of linear dynamic systems in the presence of Gaussian distribution and Gaussian noises. For nonlinear problems, the extended Kalman filter (EKF) is used. The EKF linearizes the system and measurement models using a first-order partial derivative (Jacobian) matrix while applying the KF propagation step (or the update step) to the error-state covariance matrix \cite{barShalomappandtrack2004}. The linearization procedure introduces errors in the posterior mean and covariance, leading to sub-optimal performance and at times divergence of the EKF \cite{WanUkfnonlin2000}. To circumvent the problem,  Julier and Uhlmann \cite{JUlUKF1997} introduced the unscented Kalman filter (UKF). The underlying idea was that a probability distribution is relatively easier to approximate than an arbitrary nonlinear function. To this end, UKF uses carefully chosen, sigma points that capture the state mean and covariance. When propagated through the nonlinear system, it also captures the posterior mean and covariance accurately to the third order \cite{WanUkfnonlin2000}. \\
Generally, the UKF is more accurate than the EKF and obviates the need to calculate the Jacobian matrix. Like the EKF, it requires the prior statistical characteristic of system noise, process noise covariance, and the measurement noise covariance to be precisely known, specifically because the covariances directly regulate the effect of prediction values and measurements on system state estimations. Describing the exact noise covariance, which may change during operation, is a challenging task but an important one because uncertainty may reduce the filter estimation performance or even lead to its divergence.  Therefore, methods for adjusting the process and measurement noise covariance matrices have been suggested in the literature, introducing the concept of adaptive filters. \\
A common approach to adaptive filtering is covariance matching \cite{covmatchservey2010}. This method uses the innovation (difference between the actual measurement and its predicted prior value) and residual (difference between the real measurement and its estimated posterior value) vectors to construct a noise statistics estimator to estimate and tune the process and measurement covariance matrices in real time.
A robust adaptive UKF (RAUKF) was introduced \cite{zhadapUkf2018} to adaptively adjust the noise covariance matrices according to the current estimation and the previous value only when a statistical fault is detected. In \cite{ChasistUKF2020}, the algorithm used to adjust the process and measurement noise covariance matrices is based on the residual and innovation sequences, using a moving window to improve the navigation performance of an autonomous underwater vehicle. Correlation and covariance matching play a role also in \cite{JwfuzzaUKF2010}, where the authors adjust the error covariance, process, and measurement noise covariance matrices using a scaling factor that decreases in time to obtain better performance for ultra-tight global positioning system (GPS)-integrated navigation. Covariance matching was used also in \cite{mencovmatchukf2016} to improve the performance of integrated GPS in a navigation system using the residual and innovation sequences in a moving window time frame. \\
Recently, machine learning (ML) and deep learning (DL) algorithms have been embedded into inertial sensing and sensor fusion problems (\cite{NadavkleInertNavServey2023}, \cite{ChenPanLearnInertServey2024}), including for adaptive process noise covariance estimation.  \cite{yangvaraukf2021} proposed variational Bayesian adaptive UKF for indoor localization where the adjustment of the process and measurement noise matrices are carried out by performing variational approximation, using Wishart prior distribution and the likelihood of current states and current measurements. Genetic algorithm (GA) support vector regression (SVR) (an SVR GA optimized approach) was proposed in \cite{xueaukf2020} to better optimize the UKF, based on the moving window covariance matching method. Another approach is to update the UKF process and measurement noise covariance matrices using the intuitionistic fuzzy logic method \cite{fangfuzzyaukf2022}. Subsequently, a set of papers proposed different neural network architectures to regress the process noise covariance in nonlinear navigation estimation frameworks. Initially, a hybrid learning-based adaptive EKF filter was proposed in \cite{kleinhybridekf2022}, where, a deep neural network model that tunes the momentary system noise covariance based only on the inertial
sensor readings was suggested. VIO-DualProNet is a novel DL method to dynamically estimate the inertial noise and integrate it into the visual-inertial navigation system (VINS-Mono) algorithm \cite{SolodarKleinVioDualNet2024}, where a process noise network, ProNet, was designed separately for the accelerometer and gyroscope readings. In \cite{Nadavkleinhybridekf2024}, the authors proposed an adaptive EKF, where a set-transformer network learns the varying process noise covariance in real time, adjusting an EKF, which optimized a navigation algorithm based on an inertial navigation system (INS) and Doppler velocity log (DVL).\\
Motivated by the above adaptive hybrid EKF methods, we propose the adaptive neural UKF (ANUKF) capable of tuning the process noise matrix dynamically and thereby coping with the nonlinear characteristics of the inertial fusion problem. ANUKF uses simple yet efficient enhanced ProcessNet networks that capture the dynamic system uncertainties, feeding them through a process noise covariance matrix. To demonstrate our ANUKF, we adopted the INS/DVL fusion for autonomous underwater vehicles (AUVs). Using a real recorded dataset from an AUV, we compared our approach with the standard UKF (fixed process noise covariance) and an adaptive neural EKF (ANEKF) that uses the same enhanced ProcessNet networks, creating a fair comparison. 
We show that our ANUKF outperforms both approaches in normal DVL conditions and in situations of DVL outages.\\
The rest of the paper is organized as follows: Section \ref{prob_form_sec} describes the UKF algorithm.
Section \ref{prop_appr} introduces our proposed approach including the neural network architecture, its training procedure, and the process noise matrix updating procedure.
Section \ref{res_sec} presents our AUV recorded dataset and results. Finally, Section \ref{conc_sec} provides the conclusions of this work.

\section{Problem Formulation}\label{prob_form_sec}
\subsection{The UKF algorithm}\label{ukf_std_alg}
As noted above, the process noise Q and measurement noise R matrices may change because of dynamic or environmental conditions. Incorrect estimation of these matrices may cause inaccurate estimation of the UKF or even its divergence.
Consider UKF to estimate the unobserved state vector {x} of the following dynamic system:
\begin{equation}\label{general_dynamic_model_eq}
    \centering
    \boldsymbol{x}_{k+1} = {f}(\boldsymbol{x}_{k}, \boldsymbol{\omega}_{k})
\end{equation}
where ${x}_{k}$ is the unobserved state at step ${k}$, ${f}$ is the dynamic model mapping, and ${\omega}$ is the process noise.
\begin{equation}\label{general_meas_model_eq}
    \centering
    \boldsymbol{z}_{k+1} = {h}(\boldsymbol{x}_{k+1}, \boldsymbol{\nu}_{k+1})
\end{equation}
where, ${h}$ is the observation model mapping, ${\nu}$ is the measurement noise, and ${z_{k+1}}$ is the observed signal at step ${k+1}$.
UKF \cite{JUlUKF1997} is based on the scaled unscented transform (UT), a method for calculating the statistics of a random variable that undergoes a nonlinear transformation. It starts with a selected population, propagates the population through the nonlinear function, and computes the appropriate probability distribution (mean and covariance) of the new population. The population is carefully selected to induct the mean and covariance of the undergoing population by the nonlinear transformation.
The UKF is an iterative algorithm. The initialization step is followed by iterative cycles, each consisting of several steps. Step 1 is the initialization, steps 2-4 are executed iteratively (step 4 upon measurement availability). These steps are described below.

\begin{enumerate}
    \item Initialization\\
        Given the initializing random ${n}$ state vector ${x}_{0}$ with known mean and covariance, set
        \begin{equation}\label{init_x_eq}
            \centering
            \hat{\boldsymbol{x}}_{0} = {E[{x}_{0}]}
        \end{equation}
        \begin{equation}\label{init_p_eq}
            \centering
            \boldsymbol{\mathbf{P}}_{0} = {E[ ({x}_{0}-\hat{\boldsymbol{x}}_{0})({x}_{0}-\hat{\boldsymbol{x}}_{0})^{T} ]}
        \end{equation}
        set the weights for the mean (uppercase letters m) computation and the covariance (uppercase letters c):
         \begin{equation}\label{w_m_0_eq}
            \centering
            \mathbf{w}_{0}^{m} = \frac{\lambda}{(n+\lambda)}
        \end{equation}
         \begin{equation}\label{w_c_0_eq}
            \centering
            \mathbf{w}_{0}^{c} = \frac{\lambda}{(n+\lambda)} + {(1 + \alpha^2 + \beta)}
        \end{equation}
         \begin{equation}\label{w_i_eq}
            \centering
            \mathbf{w}_{i}^{m} = \mathbf{w}_{i}^{c} = \frac{1}{2(n+\lambda)}, {i=1,...,2n}
        \end{equation}
            
        \item Update sigma points\\
        Set ${2n+1}$ sigma points as follows:
         \begin{equation}\label{sp_0_eq}
            \centering
            \boldsymbol{x}_{0,{{k}{|}{k}}} = \hat{\boldsymbol{x}}_{{k}{|}{k}}
        \end{equation}
         \begin{equation}\label{sp_i_eq}
            \centering
            \begin{aligned}
            \boldsymbol{x}_{i,{{k}{|}{k}}} =& \hat{\boldsymbol{x}}_{{k}{|}{k}} + (\sqrt{({n}+\lambda)\boldsymbol{\mathbf{P}}_{{k}{|}{k}}})_{i},
            \\& {{i}= {1},...,{n}}
            \end{aligned}
       \end{equation}
        \begin{equation}\label{sp_n_i_eq}
            \begin{aligned}
            \centering
            \boldsymbol{x}_{i,{{k}{|}{k}}} =& \hat{{x}}_{{k}{|}{k}} - (\sqrt{({n}+\lambda)\boldsymbol{\mathbf{P}}_{{k}{|}{k}}})_{i-n}, \\
            & {{i}= {n+1},...,{2n}}
            \end{aligned}
        \end{equation}
        where $\boldsymbol{\mathbf{P}}_{{k}{|}{k}}$ is positive definite, therefore it can be factored by Cholesky decomposition as $\sqrt{\boldsymbol{\mathbf{P}}_{{k}{|}{k}}}\sqrt{\boldsymbol{\mathbf{P}}_{{k}{|}{k}}}^{T}$.
        $\sqrt{\boldsymbol{\mathbf{P}}_{{k}{|}{k}}}_i$ is the ${i}$ row of the matrix $\sqrt{\boldsymbol{\mathbf{P}}_{{k}{|}{k}}}$.
        The points are symmetrical around the expected value, so the expectation is preserved.
        ${\lambda=\alpha^2(n+\kappa)-n}$ is a scaling parameter, $\alpha$ determines the spread of the sigma points around $\hat{{x}}_{{k}{|}{k}}$, usually set to a small value ($1e-3$), $\kappa$ is a secondary scaling parameter usually set to 0, and $\beta$ is used to incorporate prior knowledge of the distribution of $x$ (for Gaussian distribution it is set to 2).        
    \item Time update\\
        The sigma points (vectors) propagate through the dynamic model mapping, after which we compute the reflected mean and covariance as follows:
        \begin{equation}\label{tu_x_eq}
            \centering
            \boldsymbol{x}_{i,{{k+1}{|}{k}}} = f(\boldsymbol{x}_{i,{{k+1}{|}{k}}}), i=0,...,2n
        \end{equation}
        \begin{equation}\label{tu_x_mean_eq}
            \centering
            \hat{\boldsymbol{x}}_{{k+1}{|}{k}} = \sum_{i=0}^{2n} \mathbf{w}_{i}^{m} \boldsymbol{x}_{i,{{k+1}{|}{k}}}
        \end{equation}
        \begin{equation}\label{tu_p_eq}
            \centering
            \boldsymbol{\mathbf{P}}_{{k+1}{|}{k}}=
            \sum_{i=0}^{2n}\mathbf{w}_{i}^{c}{\delta\boldsymbol{x}_{i,{{k+1}{|}{k}}} \cdot {\delta\boldsymbol{x}_{i,{{k+1}{|}{k}}}}^T} + \boldsymbol{\mathbf{Q}}_{k+1}
        \end{equation}
        where $\delta\boldsymbol{x}_{i,{{k+1}{|}{k}}} = \boldsymbol{x}_{i,{{k+1}{|}{k}}}-\hat{\boldsymbol{x}}_{{k+1}{|}{k}}$, $\boldsymbol{\mathbf{Q}}_{k+1}$ is the covariance process noise matrix at step ${k+1}$

    \item Measurement update\\
        Update the sigma points according to the time update phase to estimate the observation/measurement:
        \begin{equation}\label{sp_0_meas_eq}
            \centering
            \boldsymbol{x}_{0,{{k+1}{|}{k}}} = \hat{{x}}_{{k+1}{|}{k}}
        \end{equation}
        \begin{equation}\label{sp_i_meas_eq}
            \centering
            \begin{aligned}
            \boldsymbol{x}_{i,{{k+1}{|}{k}}} =& \hat{\boldsymbol{x}}_{{k+1}{|}{k}} + (\sqrt{({n}+\lambda)\boldsymbol{\mathbf{P}}_{{k+1}{|}{k}}})_{i},
            \\& {{i}= {1},...,{n}}
            \end{aligned}
        \end{equation}
        \begin{equation}\label{sp_n_i_meas_eq}
            \centering
            \begin{aligned}
            \boldsymbol{x}_{i,{{k+1}{|}{k}}} =& \hat{\boldsymbol{x}}_{{k+1}{|}{k}} - (\sqrt{({n}+\lambda)\boldsymbol{\mathbf{P}}_{{k+1}{|}{k}}})_{i-n},\\& {{i}= {n+1},...,{2n}}
            \end{aligned}
        \end{equation}
        Compute the estimated measurement of each sigma point:
        \begin{equation}\label{z_i_eq}
            \centering
            \boldsymbol{z}_{i,{{k+1}{|}{k}}} = h(\boldsymbol{x}_{i,{{k+1}{|}{k}}}), i=0,...,2n 
        \end{equation}
        Compute the mean estimated measurement:
        \begin{equation}\label{z_mean_eq}
            \centering
            \hat{\boldsymbol{z}}_{{k+1}{|}{k}} = \sum_{i=0}^{2n}{w}_{i}^{m}\boldsymbol{z}_{i,{{k+1}{|}{k}}}
        \end{equation}
        Compute the measurement covariance matrix:
        \begin{equation}\label{s_eq}
            \centering
            \boldsymbol{\mathbf{S}}_{k+1} = \sum_{i=0}^{2n}{w}_{i}^{c}\delta\boldsymbol{z}_{i,{{k+1}{|}{k}}}\cdot {\delta\boldsymbol{z}_{i,{{k+1}{|}{k}}}}^T+\boldsymbol{\mathbf{R}}_{k+1}
        \end{equation}
        where $\delta\boldsymbol{z}_{i,{{k+1}{|}{k}}} = \boldsymbol{z}_{i,{{k+1}{|}{k}}} - \hat{\boldsymbol{z}}_{{k+1}{|}{k}}$, $\boldsymbol{\mathbf{R}}_{k+1}$ is the measurement process noise at step ${k+1}$.
        
        Compute the cross-covariance matrix:
        \begin{equation}\label{p_cross_cov_eq}
            \centering
            \boldsymbol{\mathbf{P}}_{k+1}^{x,z} = \sum_{i=0}^{2n}{w}_{i}^{c}\delta\boldsymbol{x}_{i,{{k+1}{|} {k}}}\cdot \delta\boldsymbol{z}_{i,{{k+1}{|}{k}}}^T
        \end{equation}
        where $\delta\boldsymbol{x}_{i,{{k+1}{|}{k}}} = \boldsymbol{x}_{i,{{k+1}{|}{k}}}-\hat{\boldsymbol{x}}_{{k+1}{|}{k}}$ and $\delta\boldsymbol{z}_{i,{{k+1}{|}{k}}} = \boldsymbol{z}_{i,{{k+1}{|}{k}}} - \hat{\boldsymbol{z}}_{{k+1}{|}{k}}$.\\
        Compute the Kalman gain:
        \begin{equation}\label{k_eq}
            \centering
            \boldsymbol{\mathbf{K}}_{k+1} = \boldsymbol{\mathbf{P}}_{k+1}^{x,z}\boldsymbol{\mathbf{S}}_{k+1}^{-1}
        \end{equation}    
        Compute the new mean according to the observation $z_{k+1}$:
        \begin{equation}\label{x_meas_eq}
            \centering
            \hat{\boldsymbol{x}}_{{k+1}{|}{k+1}}=\hat{\boldsymbol{x}}_{{k}{|}{k+1}} + \boldsymbol{\mathbf{K}}_{k+1}(z_{k+1}-\hat{\boldsymbol{z}}_{{k+1}{|}{k}})
        \end{equation}    
        Compute the new covariance matrix using the Kalman gain:
        \begin{equation}\label{P_mean_eq}
            \centering
            \boldsymbol{\mathbf{P}}_{{k+1}{|}{k+1}} = \boldsymbol{\mathbf{P}}_{{k+1}{|}{k}}-\boldsymbol{\mathbf{K}}_{k+1}\boldsymbol{\mathbf{S}}_{k+1}{\boldsymbol{\mathbf{K}}_{k+1}}^{T} 
        \end{equation}
\end{enumerate}
    Equation \eqref{tu_p_eq} introduces $\boldsymbol{\mathbf{Q}}_{k+1}$, the process noise covariance matrix at step ${k+1}$, affecting directly  $\boldsymbol{\mathbf{P}}_{{k+1}{|}{k}}$, the error covariance matrix. As noted, the process noise may change. It is critical to estimate it correctly, otherwise loss of accuracy or even divergence may occur.

\section{Proposed Approach}\label{prop_appr}
\begin{figure}[h]
    \centering
    \includegraphics[width=0.9\linewidth]{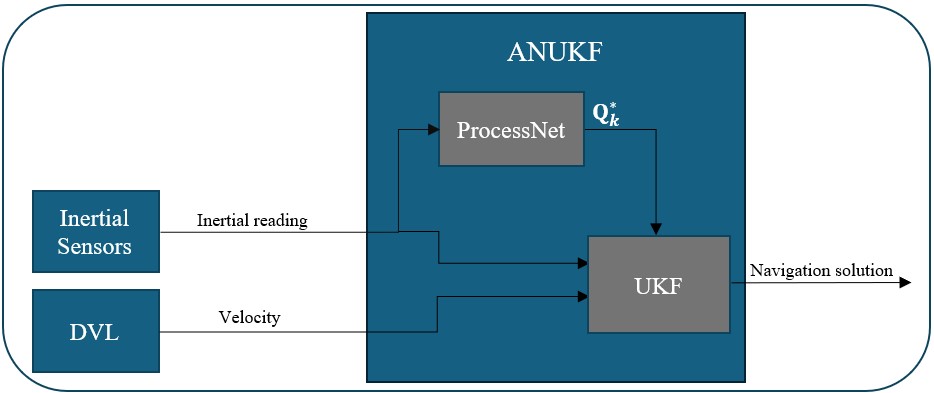}
    \caption{Our proposed ANUKF implemented on the DVL/INS fusion problem.}
    \label{fig:DNN-AUKF-Diagram}
\end{figure}

To cope with real-time adaptive process noise covariance matrix estimation in nonlinear filtering, we propose ANUKF, an adaptive neural unscented Kalman filter. The motivation for our approach stems from the fact that although the UKF structure enables it to better handle nonlinear problems, it lacks the ability to cope with real-world varying uncertainty. By contrast, our ANUKF offers a simple and efficient neural network to learn the uncertainty of the process noise covariance using only inertial sensor readings. For demonstration purposes, we adopted the DVL and INS (DVL/INS) fusion problem but the proposed ANUKF can be applied to any inertial fusion problem.\\
A block diagram of ANUKF for DVL/INS fusion is presented in Figure \ref{fig:DNN-AUKF-Diagram}. The inertial sensor readings and the estimated velocity of the DVL are plugged into the UKF. Additional inputs to our regression network, ProcessNet, are the inertial readings. ANUKF outputs the adaptive process noise covariance matrix required by the UKF mechanism and gives the full estimated navigation solution. 

\subsection{ANUKF for DVL/INS fusion}\label{adnn_ukf_alg}
The following 12 error state vector is used in the DVL/INS settings:
\begin{equation}\label{err_state_eq}
    \centering
    \delta{\boldsymbol{x}}=\begin{bmatrix}
    \delta{\boldsymbol{v}^n} & \boldsymbol\delta{\Psi}^n & \boldsymbol{b}_a & \boldsymbol{b}_g
    \end{bmatrix} \in \mathbb{R}^{12 \times 1}
\end{equation}
where $\delta{\boldsymbol{v}^n}\in\mathbb{R}^3$ is the velocity error state vector, expressed in the navigation frame,
$\boldsymbol\delta\Psi^n \in \mathbb{R}^3$ is the misalignment error state vector, expressed in the navigation frame,
$\boldsymbol{b}_a\in\mathbb{R}^3$ is the accelerometers residual bias vector expressed in the body frame, and $\boldsymbol{b}_g\in\mathbb{R}^3$ is the gyroscope's residual bias vector expressed in the body frame. The body frame is an orthogonal axis set (x,y,z), centered on the reference point of the inertial measurements: the x-axis aligns with the AUV’s longitudinal axis, pointing in the forward direction; the z-axis extends downward; and the y-axis extends outward, completing the right-hand orthogonal coordinate system.
The inertial measurements are modeled with their true values, with the addition of bias and zero mean white Gaussian noise:
\begin{equation}\label{acc_noise_meas}
    \centering
    \boldsymbol{f}_{ib}^b = \boldsymbol{f}_{ib,t}^b + \boldsymbol{b}_a + \boldsymbol{n}_a
\end{equation}
\begin{equation}\label{gyro_noise_meas}
    \centering
    \boldsymbol{\omega}_{ib}^b = \boldsymbol{\omega}_{ib,t}^b + \boldsymbol{b}_g + \boldsymbol{n}_g
\end{equation}
where $f_{ib,t}^b$ is the measured specific force vector, $\boldsymbol{\omega}_{ib,t}^b$ is the measured angular velocity expressed in the body frame, $b_i$ $i=a,g$ is the bias vector of the accelerometer and gyroscope, respectively, and $n_i$ $i=a,g$ is the zero mean white Gaussian noise of the accelerometer and gyroscope, respectively.
The biases are modeled as random walk processes:
\begin{equation}\label{acc_bias_rw}
    \centering
    \dot{\boldsymbol{b}}_a = \boldsymbol{n}_{a_b} \sim\mathcal{N}(0,{\sigma_{\boldsymbol{a}_b}}^2)
\end{equation}
\begin{equation}\label{gyro_bias_rw}
    \centering
    \dot{\boldsymbol{b}}_g = \boldsymbol{n}_{g_b} \sim \mathcal{N}(0,{\sigma_{\boldsymbol{g}_b}}^2)
\end{equation}
where ${\sigma_{\boldsymbol{a}_b}}$ and ${\sigma_{\boldsymbol{g}_b}}$ are the accelerometer and gyroscope Gaussian noise standard deviations, respectively.
The UKF error state continuous time model is\\
\begin{equation}\label{x_dot_eq}
    \centering
    \dot\delta{\boldsymbol{x}} = \boldsymbol{\mathbf{F}}\delta{\boldsymbol{x}} + \boldsymbol{\mathbf{G}}\delta{\boldsymbol{w}}
\end{equation}    
where $\boldsymbol{\mathbf{F}}\in\mathbb{R}^{12 \times 12}$ is the system matrix (see \cite{GrovePrincipleGNSS} section 12.2.4), $\delta{\boldsymbol{w}}=[\boldsymbol{n}_a^T  \boldsymbol{n}_g^T  {{\boldsymbol{n}_a}_b}^T  {{\boldsymbol{n}_g}_b}^T]^T \in \mathbb{R}^{12 \times 1}$ is the system noise vector, and $G \in \mathbb{R}^{12 \times 12}$ is the system noise distribution matrix
\begin{equation}\label{G_mat_eq}
    \centering
    \boldsymbol{\mathbf{G}}=\begin{bmatrix}
    {\boldsymbol{\mathbf{C}}_b^n} & \boldsymbol{\mathbf{0}}_{3 \times 3} & \boldsymbol{\mathbf{0}}_{3 \times 3} & \boldsymbol{\mathbf{0}}_{3 \times 3} \\
    \boldsymbol{\mathbf{0}}_{3 \times 3} & {\boldsymbol{\mathbf{C}}_b^n} & \boldsymbol{\mathbf{0}}_{3 \times 3} & \boldsymbol{\mathbf{0}}_{3 \times 3}\\
    \boldsymbol{\mathbf{0}}_{3 \times 3} & \boldsymbol{\mathbf{0}}_{3 \times 3} & \boldsymbol{\mathbf{I}}_{3\times3} & \boldsymbol{\mathbf{0}}_{3 \times 3}\\ 
    \boldsymbol{\mathbf{0}}_{3 \times 3} & \boldsymbol{\mathbf{0}}_{3 \times 3} & \boldsymbol{\mathbf{0}}_{3 \times 3} & \boldsymbol{\mathbf{I}}_{3\times3} 
    \end{bmatrix}
\end{equation}    
with ${\boldsymbol{\mathbf{C}}_b^n}$ representing the rotation matrix from the body frame to the navigation frame, as calculated by the navigation algorithm, $\boldsymbol{\mathbf{0}}_{3 \times 3}$ is the zero matrix, and $\boldsymbol{\mathbf{I}}_{3 \times 3}$ is the identity matrix.\\
The discrete process noise covariance is obtained based on the continuous process noise covariance and given by (\cite{barShalomappandtrack2004} section 4.3.4): 
\begin{equation}\label{Q_calc_eq}
    \centering
    \boldsymbol{\mathbf{Q}}_{k+1} =  \boldsymbol{\mathbf{G}}_{k+1} \cdot \boldsymbol{\mathbf{Q}}^*_{k+1} \cdot \boldsymbol{\mathbf{G}}_{k+1}^T
\end{equation}    
where 
\begin{equation}\label{Q_star_mat_eq}
    \centering
    \boldsymbol{\mathbf{Q}}^{*}_{k+1} = diag(\delta\omega)\in \mathbf{R}^{12x12}
\end{equation}\\
The measurement, \eqref{x_meas_eq}, requires the calculation of the mean estimated measurement \eqref{z_mean_eq}. It depends on $\boldsymbol{z}_{i,{{k+1}{|}{k}}}$, the estimation of the velocity error vector in the body frame axes of sigma-point ${i}$, $i=0,...,2n$, as shown in \eqref{z_i_eq}. To this end, we computed $\boldsymbol{z}_{i,{{k+1}{|}{k}}}$ as follows:
\begin{equation}\label{ukf_dvl_meas_func}
    \begin{split}
    \boldsymbol{z}_{i,{{k+1}{|}{k}}} = h(\boldsymbol{x}_{i,{{k+1}{|}{k}}}) = \boldsymbol{\mathbf{C}}_{n}^{b}\cdot \hat{\boldsymbol{\mathbf{C}}}_{n_i}^{n} & [\boldsymbol{v}_n + \Delta \hat{\boldsymbol{v}}_{n_i}] - \boldsymbol{\mathbf{C}}_{n}^{b}\cdot\boldsymbol{v}_n,\\
    & i=0,...,2n
    \end{split}
\end{equation}
where, $C_{n}^{b}$ is the computed rotation matrix from the navigation frame to the AUV body frame, $\hat{C}_{n_i}^{n}$ is the estimated misalignment ($\delta{\Psi_i^n}$) of sigma point $i$, $\boldsymbol{v}_n$ is the computed velocity in the navigation frame, and $\Delta \hat{\boldsymbol{v}}_{n_i}$ is the estimated velocity error of sigma point $i$ given in the navigation frame.

\subsection{ProcessNet Structure}\label{adnn_alg}
\begin{figure*}[h]
    \centering
    \includegraphics[width=\textwidth]{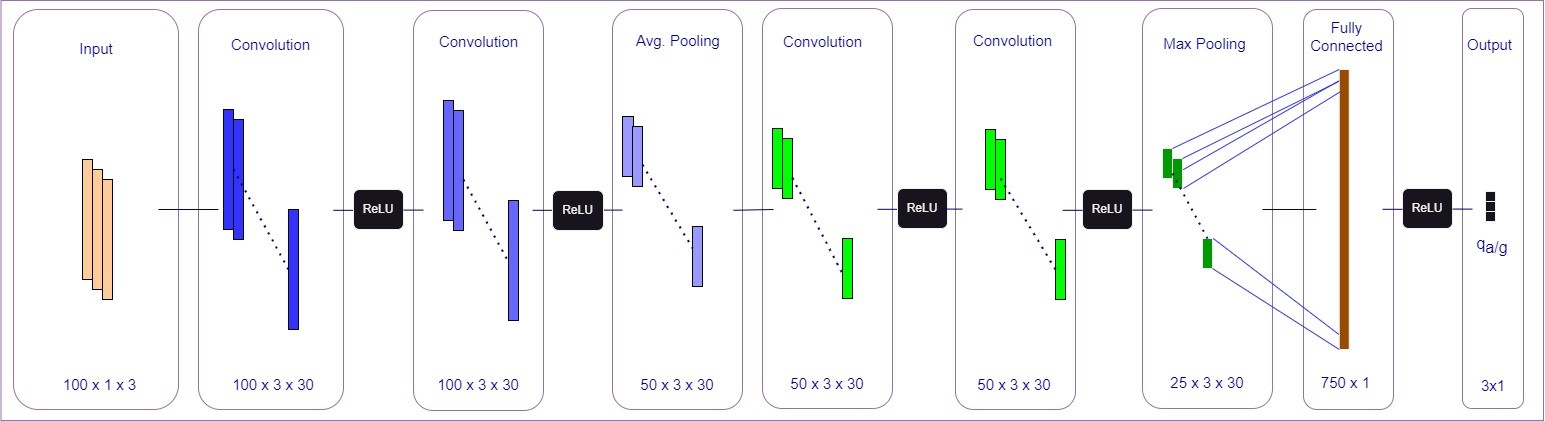}
    \caption{ProcessNet: the accelerometer and gyroscope process noise covariance regression network structure.}
    \label{fig:Dynam-Q-acc-CNN-Diagram}
\end{figure*}

The dynamic accelerometers and gyroscopes process noise covariance network architecture is described in Figure \ref{fig:Dynam-Q-acc-CNN-Diagram}. Although the network structure is identical, the accelerometer ProcessNet and gyroscope ProcessNet differ in their input and output. The input to the accelerometer ProcessNet network at time step $k$ is the accelerometer readings in a predefined window size and the output is the diagonal entries in the process noise covariance matrix that corresponds to the accelerometers.
The dynamic gyroscope ProcessNet network architecture is similar to the accelerometer network but it has scale components to the input and the output, forcing the network to work with sufficiently large numbers. The network inputs at time step $k$ are the gyroscope readings in a predefined window size and the output is the diagonal entries in the process noise covariance matrix that corresponds to the gyroscopes.\\
Next, we defined the network layers, as shown in Figure \ref{fig:Dynam-Q-acc-CNN-Diagram}. The network receives an input $d_{in} \in \mathbb{R}^{w \times 3}$, where 3 is the number of input channels and $w$ is the window size length (number of samples in each input). Each layer is denoted by $l_i$, where $i$ is the number of the layer, $\boldsymbol{b}_i$ is the bias vector, and ${\boldsymbol{\mathbf{\Theta}}_i}$ represents the weights of layer $i$. We also used the rectified linear unit (ReLU) activation function between each layer defined by \eqref{ReLU_form},  followed by max/average pooling.
\begin{equation}\label{ReLU_form}
    \centering
    ReLU(x) = max(0,x)
\end{equation}
where the operation is defined elementwise (if $x\in \mathbb{R}^{m \times n}$ then also $ReLU(x)\in \mathbb{R}^{m \times n}$).
Using a window size of $w=100$, the first layer is defined by:
\begin{equation}\label{Q_net_l1}
   \begin{aligned}
    \boldsymbol{\mathbf{l}}_1(i,k)= & max([\sum_{j=1}^{3}{\boldsymbol{\mathbf{d}}_{in}}_{pad} (i:i+2,j)\\ & *\boldsymbol{\mathbf{\Theta}}_1(j,k,1:3)^T]+\boldsymbol{b}_1(k),0)\\ &
    i=1,...100, k=1,...,30.
   \end{aligned}
\end{equation}
where ${\boldsymbol{\mathbf{d}}_{in}}_{pad} \in \mathbb{R}^{102 \times 3}$ is the padded input, ${\boldsymbol{\mathbf{d}}_{in}}_{pad} (i:i+2,j)$ is the row vector $[{\boldsymbol{\mathbf{d}}_{in}}_{pad} (i,j), {\boldsymbol{\mathbf{d}}_{in}}_{pad} (i+1,j), {\boldsymbol{\mathbf{d}}_{in}}_{pad} (i+2,j)]$, $*$ is the cross-correlation operator (if $\boldsymbol{a}, \boldsymbol{c} \in  \mathbb{R}^{3 \times 1}$, $\boldsymbol{a}^T * \boldsymbol{c} = \sum_{h=1}^{3}\boldsymbol{a}(h) \cdot \boldsymbol{c}(h)$).
${\boldsymbol{l}_1} \in \mathbb{R}^{100 \times 30}$, ${\boldsymbol{\mathbf{\Theta}}_1} \in \mathbb{R}^{3 \times 30 \times 3}$, ${\boldsymbol{b}_1} \in \mathbb{R}^{30 \times 1}$, and $max$ is the ReLU operation.\\
The second layer is defined by:
\begin{equation}\label{Q_net_l2}
   \begin{aligned}
    \boldsymbol{\mathbf{l}}_2(i,k)= & max([\sum_{j=1}^{30}{\boldsymbol{\mathbf{l}}_1}_{pad}(i:i+2,j) \\ & *\boldsymbol{\mathbf{\Theta}}_2(j,k,1:3)^T]+\boldsymbol{b}_2(k),0)\\ &
    i=1,...100, k=1,...,30.
   \end{aligned}
\end{equation}
with ${\boldsymbol{\mathbf{l}}_2} \in \mathbb{R}^{100 \times 30}$, ${\boldsymbol{\mathbf{l}}_1}_{pad} \in \mathbb{R}^{102 \times 30}$, ${\boldsymbol{\mathbf{\Theta}}_2} \in \mathbb{R}^{30 \times 30 \times 3}$, and $\boldsymbol{b}_2 \in \mathbb{R}^{30 \times 1}$.\\
The third layer, ${\boldsymbol{\mathbf{l}}_3}\in \mathbb{R}^{50 \times 30}$, is:
\begin{equation}\label{Q_net_l3}
   \begin{aligned}
    \centering
    \boldsymbol{\mathbf{l}}_3(i,k) = & 0.5\cdot {\boldsymbol{\mathbf{l}}_2}(2\cdot i-1,k) + {\boldsymbol{\mathbf{l}}_2}(2\cdot i,k)\\
    & i=1,...50, k=1,...,30.
    \end{aligned}
\end{equation}
and the forth layer, ${\boldsymbol{\mathbf{l}}_4} \in \mathbb{R}^{50 \times 30}$, is:
\begin{equation}\label{Q_net_l4}
   \begin{aligned}
    \centering
    \boldsymbol{\mathbf{l}}_4(i,k) = & max([\sum_{j=1}^{30}{\boldsymbol{\mathbf{l}}_3}_{pad}(i:i+2,j) \\ & *\boldsymbol{\mathbf{\Theta}}_3(j,k,1:3)^T]+\boldsymbol{b}_3(k),0) \\ 
    & i=1,...50, k=1,...,30.
    \end{aligned}
\end{equation}
where ${\boldsymbol{l}_3}_{pad} \in \mathbb{R}^{52 \times 30}$, ${\boldsymbol{\mathbf{\Theta}}_3} \in \mathbb{R}^{30 \times 30 \times 3}$, and ${\boldsymbol{b}_3} \in \mathbb{R}^{30 \times 1}$. \\
The fifth layer, ${\boldsymbol{\mathbf{l}}_5} \in \mathbb{R}^{50 \times 30}$, is defined by:
\begin{equation}\label{Q_net_l5}
   \begin{aligned}
    \centering
    \boldsymbol{\mathbf{l}}_5(i,k) = & max([\sum_{j=1}^{30}{\boldsymbol{\mathbf{l}}_4}_{pad}(i:i+2,j) \\ & *\boldsymbol{\mathbf{\Theta}}_4(j,k,1:3)^T]+\boldsymbol{b}_4(k),0) \\
    & i=1,...50, k=1,...,30.
    \end{aligned}
\end{equation}
where ${\boldsymbol{\mathbf{l}}_4}_{pad} \in \mathbb{R}^{52 \times 30}$, ${\boldsymbol{\mathbf{\Theta}}_4} \in \mathbb{R}^{30 \times 30 \times 3}$, and ${\boldsymbol{b}_4} \in \mathbb{R}^{30 \times 1}$.. \\
 The sixth and last convolution layer, ${\boldsymbol{\mathbf{l}}_6} \in \mathbb{R}^{25 \times 30}$, is:
\begin{equation}\label{Q_net_l6}
   \begin{aligned}
    \centering
    \boldsymbol{\mathbf{l}}_6(i,k) = & max({\boldsymbol{\mathbf{l}}_5}(2\cdot i-1,k),{\boldsymbol{\mathbf{l}}_5}(2\cdot i,k)) \\
    & i=1,...25, k=1,...,30.
   \end{aligned}
\end{equation}
The next operation is to flatten $\boldsymbol{\mathbf{l}}_6$ to form ${\boldsymbol{l}_7} \in \mathbb{R}^{750 \times 1}$.
The last hidden layer is a fully connected one with a linear operation:
\begin{equation}\label{Q_net_res}
    \centering
    \boldsymbol{net_{res}} = [{\boldsymbol{l}_{7}^{T}} \cdot \boldsymbol{\mathbf{\Theta}}_5]^T + \boldsymbol{b}_5
\end{equation}
where ${\boldsymbol{\mathbf{\Theta}}_5} \in \mathbb{R}^{750 \times 3}$ and ${\boldsymbol{b}_5} \in \mathbb{R}^{3 \times 1}$, so that ${\boldsymbol{net_{res}}} \in \mathbb{R}^{3 \times 1}$.\\

\subsection{Training Process}\label{adnn_train}
The mean square error (MSE) loss function is used in both the accelerometer and gyroscope ProcessNets. The loss is defined by:
\begin{equation}\label{Q_net_cost}
    \centering
     \mathbf{l}_{MSE}(\mathbf{N_k},\hat{\mathbf{N}}_k) = \frac{1}{n}\sum_{i=1}^n ||\mathbf{N_{k_i}}-\hat{\mathbf{N}}_{k_i}||^2
\end{equation}
where $n$ is the batch size, $\hat{N}_k, N_k \in\mathbb{R}^{n \times 3}$ are the ProcessNet batched estimations and the ground truth (GT), with $k={a,g}$ for the accelerometer and gyroscope networks, respectively.

\subsection{Updating the ANUKF process noise}\label{adnn_updt}
The general structure of the regressed adaptive process noise matrix is:
\begin{equation}\label{Net_Q_star_mat_eq}
    \centering
    \boldsymbol{\mathbf{Q}}^{net}_{k+1}=\begin{bmatrix}
    {\boldsymbol{\mathbf{q}}_v}_{3 \times 3} & \boldsymbol{\mathbf{0}}_{3 \times 3} & \boldsymbol{\mathbf{0}}_{3 \times 3} & \boldsymbol{\mathbf{0}}_{3 \times 3} \\
    \boldsymbol{\mathbf{0}}_{3 \times 3} & {\boldsymbol{\mathbf{q}}_\Psi}_{3 \times 3} & \boldsymbol{\mathbf{0}}_{3 \times 3} & \boldsymbol{\mathbf{0}}_{3 \times 3}\\
    \boldsymbol{\mathbf{0}}_{3 \times 3} & \boldsymbol{\mathbf{0}}_{3 \times 3} & {\boldsymbol{\mathbf{q}}_a}_{3 \times 3} & \boldsymbol{\mathbf{0}}_{3 \times 3}\\ 
    \boldsymbol{\mathbf{0}}_{3 \times 3} & \boldsymbol{\mathbf{0}}_{3 \times 3} & \boldsymbol{\mathbf{0}}_{3 \times 3} & {\boldsymbol{\mathbf{q}}_g}_{3 \times 3} 
    \end{bmatrix}
\end{equation}    
Next, we define each $3 \times 1$ vector that assembles the diagonal elements of the non-zero $3 \times 3$ submatrices in \eqref{Net_Q_star_mat_eq}. The accelerometer network output $[q_{a_x}, q_{a_y}, q_{a_z}]$ is multiplied by the integration interval $\tau$, so that
\begin{equation}\label{Q_q_a_elem_eq}
    \centering
    \boldsymbol{q}_a  = [q_{a_x}, q_{a_y}, q_{a_z}] \cdot \tau
\end{equation}
Using \eqref{Q_q_a_elem_eq}, the velocity uncertainty is:
\begin{equation}\label{Q_q_v_elem_eq} 
    \centering
    \boldsymbol{q}_v = \boldsymbol{q}_a \cdot \tau_a
\end{equation}
where $\tau_a$ is a factor that depends on the inertial measurement rate.
In the same manner, the gyroscope network outputs $[q_{g_x}, q_{g_y}, q_{g_z}]$ are multiplied by the integration interval $\tau$, so that
\begin{equation}\label{Q_q_g_elem_eq}
    \centering
    \boldsymbol{q}_g = [q_{g_x}, q_{g_y}, q_{g_z}] \cdot \tau
\end{equation}
Using \eqref{Q_q_g_elem_eq}, the orientation  uncertainty is:
\begin{equation}\label{Q_q_psi_elem_eq}
    \centering
    \boldsymbol{q}_\Psi = \boldsymbol{q}_g \cdot \tau_g
\end{equation}
where $\tau_g$ is an appropriate factor that depends on the inertial measurement rate.\\
For system stability, we ensure that the network outputs are within a valid range of
values before updating the filter, that is, we incorporate a conditioned inertial perspective to validate the network output. For example, the output should be positive (covariance matrix), not too small (perfect inertial sensors), and not too large (unrealistic behavior of the inertial sensors).\\
Next, the regressed process noise matrix \eqref{Net_Q_star_mat_eq} is plugged into \eqref{Q_calc_eq} instead of \eqref{Q_star_mat_eq} to create the ANUKF adaptive process noise covariance:
\begin{equation}\label{Q_net_calc_eq}
    \centering
    \boldsymbol{\hat{\boldsymbol{\mathbf{Q}}}}^{net}_{k+1} =  \boldsymbol{\mathbf{G}}_{k+1} \cdot \boldsymbol{\mathbf{Q}}^{net}_{k+1} \cdot \boldsymbol{\mathbf{G}}_{k+1}^T
\end{equation} 
Finally, the error state covariance matrix is updated by:
\begin{equation}\label{tu_p_qu_eq}
    \centering
    \boldsymbol{\mathbf{P}}_{{k+1}{|}{k}}=
    \sum_{i=0}^{2n}{\delta\boldsymbol{x}_{i,{{k+1}{|}{k}}}\cdot{\delta\boldsymbol{x}_{i,{{k+1}{|}{k}}}}^T} + \boldsymbol{\hat{\boldsymbol{\mathbf{Q}}}}^{net}_{k+1}
\end{equation}
where $\delta\boldsymbol{x}_{i,{{k+1}{|}{k}}} = \boldsymbol{x}_{i,{{k+1}{|}{k}}}-\hat{\boldsymbol{x}}_{i,{{k+1}{|}{k}}}$.

\section{Experimental Results}\label{res_sec}
\subsection{Dataset}\label{exp_res_dataset}
\begin{figure}[h]
    \centering
    \begin{subfigure}[b]{0.48\linewidth}
        \centering
        \includegraphics[width=\linewidth, height=3.0cm]{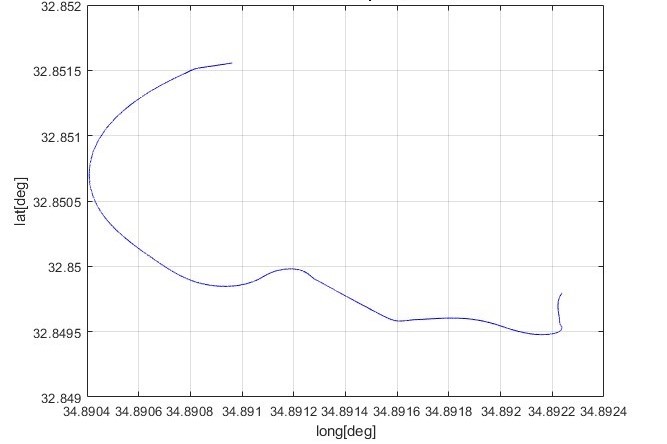}
        \caption{Track 1}
        \label{fig:track1}
    \end{subfigure}
    \hfill
    \begin{subfigure}[b]{0.48\linewidth}
        \centering
        \includegraphics[width=\linewidth]{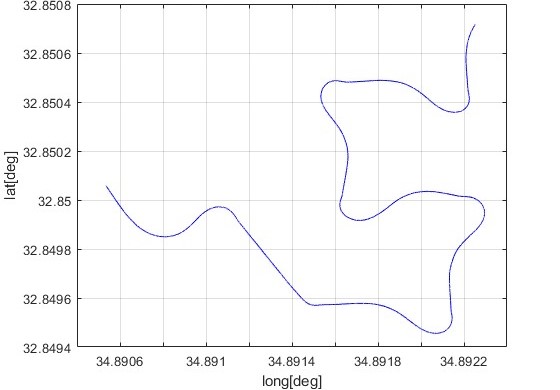}
        \caption{Track 2}
        \label{fig:track2}
    \end{subfigure}
    \vfill
    \begin{subfigure}[b]{0.48\linewidth}
        \centering
        \includegraphics[width=\linewidth]{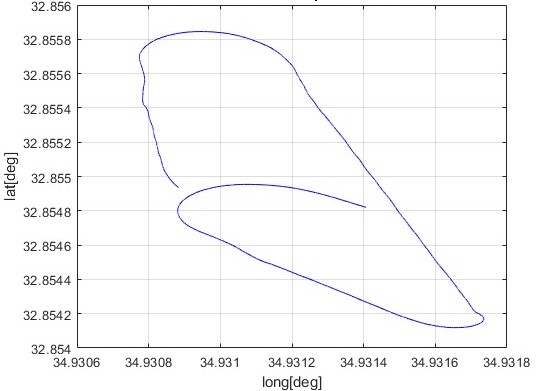}
        \caption{Track 3}
        \label{fig:track3}
    \end{subfigure}
    \hfill
    \begin{subfigure}[b]{0.48\linewidth}
        \centering
        \includegraphics[width=\linewidth]{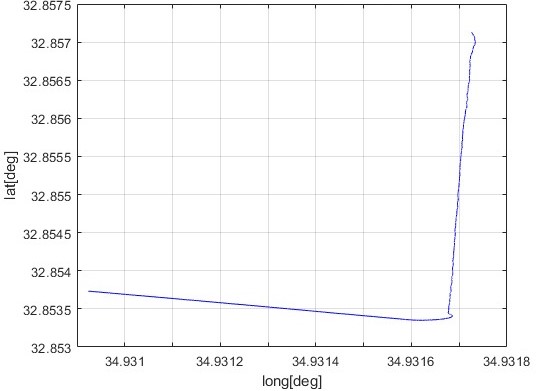}
        \caption{Track 4}
        \label{fig:track4}
    \end{subfigure}
    \caption{Horizontal position of tracks 1-4 used in the training dataset.}
    \label{fig:horiz-pos-tracks-1-4}
\end{figure}

\begin{figure}[h]
    \centering
    \begin{subfigure}[b]{0.48\linewidth}
        \centering
        \includegraphics[width=\linewidth]{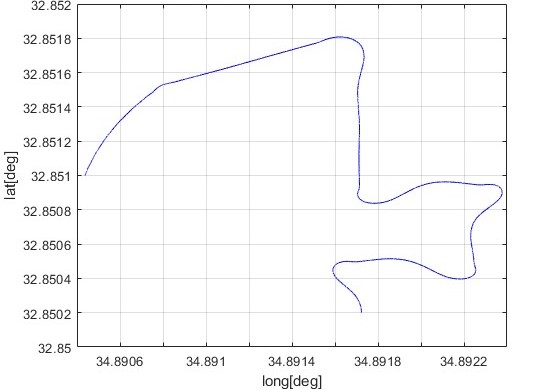}
        \caption{Track 5}
        \label{fig:track5}
    \end{subfigure}
    \hfill
    \begin{subfigure}[b]{0.48\linewidth}
        \centering
        \includegraphics[width=\linewidth]{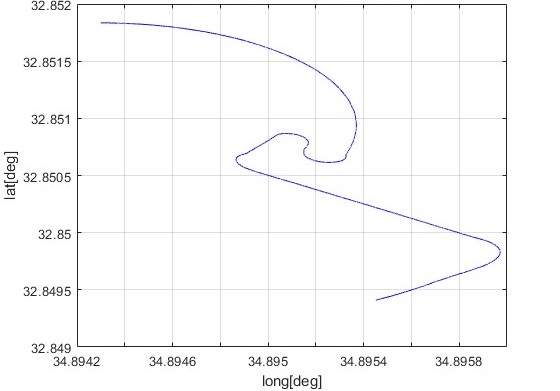}
        \caption{Track 6}
        \label{fig:track6}
    \end{subfigure}
    \caption{Horizontal position of tracks 5-6 used in the testing dataset.}
    \label{fig:horiz-pos-tracks-5-6}
\end{figure}

We used the Snapir AUV dataset (\cite{10674766}) for training and testing our proposed approach. Snapir is an ECA Robotics modified Group A18D mid-size AUV. The Snapir AUV is outfitted with the iXblue, Phins Subsea INS, which uses fiber optic gyroscope (FOG) technology for precise inertial navigation \cite{iXblue}. Snapir also uses a Teledyne RDI Work Horse Navigator DVL \cite{TeledyneMarine}, known  for its capability to provide accurate velocity measurements. The INS operates at a frequency of 100 [Hz], whereas the DVL operates at 1[Hz]. The dataset contains 24 minutes of recording divided into 6 trajectories. Each trajectory contains  the DVL/INS fusion solution (addressed as GT), inertial readings, and DVL measurements. 
In addition: 
\begin{itemize}
    \item The initial navigation solution includes Euler angles (yaw, pitch, roll) given in $[rad]$ from the body frame to the navigation frame, the velocity vector expressed in the navigation frame (north, east, down) in $[m/s]$, and the position (latitude, longitude, altitude) $[rad, rad, m]$.
    \item The IMU outputs are accelerometer and gyroscope readings expressed in the body frame (the rotation from the IMU frame to the body frame is known). The inertial sensors are sampled at 100Hz.
    \item The DVL output velocity vector $(x,y,z) [m/s]$ is expressed in the body frame (the rotation from the DVL frame to the body frame is known). The DVL measurements are sampled at 1Hz.
\end{itemize}
To create the trajectories being tested, we added the following to the given raw data: initial velocity errors with a standard deviation (STD) of $0.25[m/s]$ and $0.05[m/s]$, horizontal and vertical, respectively, and misalignment errors with a STD of $0.01[deg]$ per axis. We also added noises with a STD of $0.03[m/s^2]$ to the accelerometers and $7.3\times 10^{-6}[rad/s]$ to the gyroscopes. Finally, we added biases with a STD of $0.3[m/s^2]$ to the accelerometers and $7.3\times 10^{-5}[rad/s]$ to the gyroscopes, which have been changed dynamically (multiplied by factors of $1-6$), bringing the measurements closer to tactical grade IMU outputs.\\
For the training process, we used tracks 1-4,  each track consisting of 4 minutes, resulting in a total of 16 minutes of the training dataset. The trajectories, presented in Figure \ref{fig:horiz-pos-tracks-1-4} include a wide range of dynamics and maneuvers.
Track 1 (Figure \ref{fig:track1}) and track 2 (Figure \ref{fig:track2}) have relatively more maneuvers and dynamic changes than track 3 (Figure \ref{fig:track3}) and track 4 (Figure \ref{fig:track4}), which have longer straight sections.
The testing datasets includes tracks 5-6, presented in Figure \ref{fig:horiz-pos-tracks-5-6}, with a total time of 8 minutes. These trajectories include a wide range of dynamics and maneuvers.

\subsection{Evaluation matrices and approaches}\label{exp_res_matrices}
In \cite{Nadavkleinhybridekf2024}, the authors showed that adaptive neural EKF (ANEKF) with dynamic process noise performs better than non-adaptive EKF and adaptive EKF that uses covariance matching as the adaptation method. Therefore, we used ANEKF as the baseline and compared it with the non-adaptive UKF, and our ANUKF.\\
For a fair comparison, all three filters used the same navigation algorithm. The navigation algorithm initialized by the reference solution integrates the updated IMU outputs and reduces navigation errors by fusing the DVL measurements. Moreover, the two adaptive filters (ANEKF and ANUKF) had the same training process as our proposed network architecture: ProcessNet.\\
We carried out 100 Monte Carlo runs on the two tracks in the testing dataset.
For each track we calculated the total velocity root mean square error (VRMSE) and the total misalignment root mean square error (MRMSE) and averaged the results.
For a given track, the VRMSE is calculated as follows:
\begin{equation}\label{vrmse_calc}
    \centering
    VRMSE = \sqrt{\frac{1}{m \cdot n}\sum_{i=1}^{m}\sum_{j=1}^{n}\|{\delta \boldsymbol{v}_i (j)}\|^2}
\end{equation}
where, $n$ is the number of samples in the trajectory, $m$ is the number of MC runs, $\|{\delta \boldsymbol{v}_i (j)}\|$ is the norm of the velocity error between the estimated velocity and the GT of run $i$ in time step $j$. The track average is:
\begin{equation}\label{vrmse_avg_calc}
    \centering
    VRMSE_{avg} = \sqrt{\frac{1}{2}({VRMSE_{5}}^2 + {VRMSE_{6}}^2)}
\end{equation}
where $VRMSE_{5}$ and $VRMSE_{6}$ are the $VRMSE$ of tracks 5 and 6, respectively.\\
In the same manner, the MRMSE is:
\begin{equation}\label{Mrmse_calc}
    \centering
    MRMSE = \sqrt{\frac{1}{m\cdot n}\sum_{i=1}^{m}\sum_{j=1}^{n}\|{\delta {\boldsymbol{\Psi}}_i (j)}\|^2}
\end{equation}
where ${\delta {\boldsymbol{\Psi}}_i (j)}$ is the misalignment vector that represents by the Euler angles corresponding to the following transformation matrix $\boldsymbol{\mathbf{C}}_{n_r}^n$:
\begin{equation}\label{misaligment_err_calc}
    \centering
    \boldsymbol{\mathbf{C}}_{n_r}^n = \boldsymbol{\mathbf{C}}_b^n \cdot \boldsymbol{\mathbf{C}}_{n_r}^b
\end{equation}
and $\boldsymbol{\mathbf{C}}_{n_r}^b$ is the GT rotation matrix from the navigation frame to the body frame.

\subsection{Results}\label{exp_res_res}
\begin{figure}[h]
    \centering
    \begin{subfigure}[b]{0.48\linewidth}
        \centering
        \includegraphics[width=\linewidth]{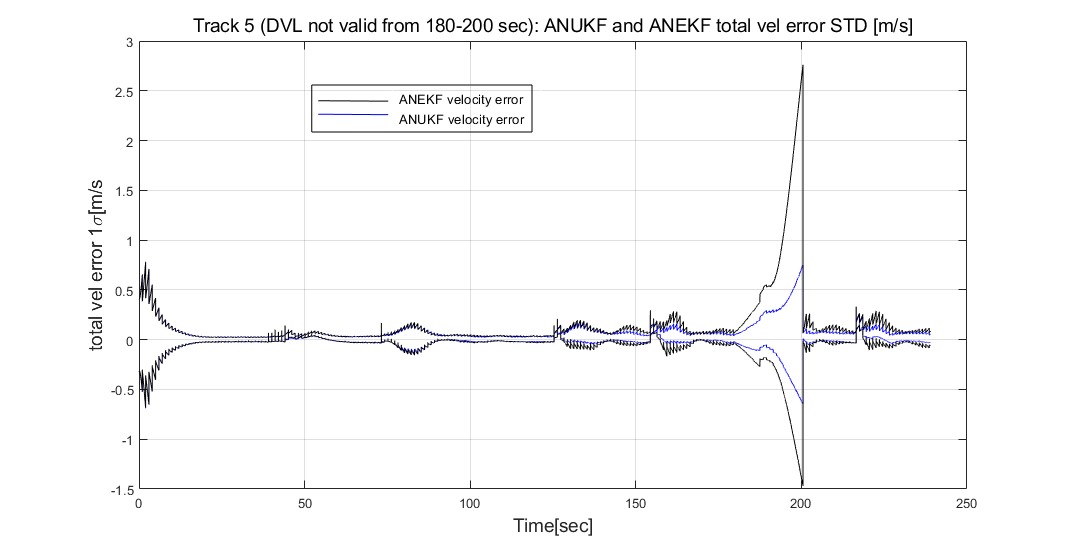}
        \caption{Velocity error track 5}
        \label{fig:vel err on track5}
    \end{subfigure}
    \hfill
    \begin{subfigure}[b]{0.48\linewidth}
        \centering
        \includegraphics[width=\linewidth]{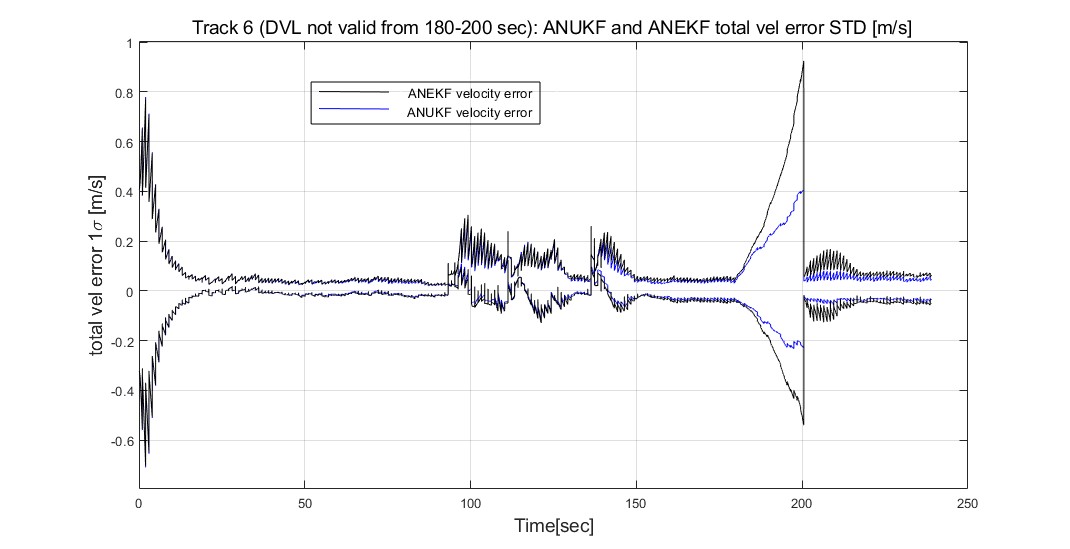}
        \caption{Velocity error track6}
        \label{fig:vel err on track6}
    \end{subfigure}
    \caption{MC velocity errors on tracks 5 and 6.}
    \label{fig:par-dvl-vel-err-tracks5-6}
\end{figure}

We applied the three filters, UKF, ANEKF, and ANUKF, to the test dataset. Table \ref{tab:dvl_full_comp_tbl} summarizes the VRMSE and MRMSE results of the three filters, showing that the velocity solution of the ANUKF performs better than the non-adaptive UKF and the ANEKF, and that the velocity solution of the ANEKF is better than that of the non-adaptive UKF. In other words, ANUKF showed an improvement of $21.3\%$ over UKF and  $8.2\%$ over ANEKF. The ANUKF estimation of misalignment showed an improvement of $5.4\%$ over UKF and $32.1\%$ over ANEKF.
\begin{table}[h]
    \centering
    \begin{tabular}{|p{1cm}|p{1cm}|p{1cm}|p{1cm}|p{1cm}|}
    \hline
    Method & VRMSE [m/sec] & ANUKF Imprv. & MRMSE [rad] & ANUKF Imprv. \\
    \hline
    \textbf{UKF} & 0.1172 & 21.3\% & 0.0112 & 5.4\% \\
    \textbf{ANEKF} & 0.1004 & 8.2\% & 0.0156 & 32.1\% \\
    \textbf{ANUKF} & 0.0922 & N/A & 0.0106 & N/A\\
    \hline
    \end{tabular}
    \caption{VRMSE and MRMSE of the UKF, ANEKF, and ANUKF applied to the testing dataset.}
    \label{tab:dvl_full_comp_tbl}
\end{table}
To further examine the robustness of our approach, we evaluated the two adaptive filters in situations of DVL outages, as is often the case in real-world scenarios \cite{10674766}. To this end, no DVL updates were provided to the filter for 20 seconds starting after 180 seconds from  starting time. We waited 180 seconds to ensure filter convergence to steady-state. Table \ref{tab:dvl_partial_comp_tbl} summarizes the VRMSE and MRMSE results of the averaging of the ANUKF and the ANEKF filters on tracks 5 and 6. The ANUKF velocity solution and misalignment estimation were significantly better than those of the ANEKF, providing a $46.7\%$ improvement in the VRMSE and $31.1\%$ in the MRMSE metric.
\begin{table}[h]
    \centering
    \begin{tabular}{|p{1cm}|p{1cm}|p{1cm}|p{1cm}|p{1cm}|}
    \hline
    Method & VRMSE [m/sec] & ANUKF Imprv. & MRMSE [rad] & ANUKF Imprv. \\
    \hline
    \textbf{ANEKF} & 0.222 & 46.7\% & 0.0156 & 32.1\% \\
    \textbf{ANUKF} & 0.1183  & N/A & 0.0106 & N/A\\
    \hline
    \end{tabular}\\
    \caption{VRMSE and MRMSE of the UKF, ANEKF, and ANUKF applied to the testing dataset. DVL outage was enforced for 20 seconds during the trajectories.}
    \label{tab:dvl_partial_comp_tbl}
\end{table}
Figure (\ref{fig:par-dvl-vel-err-tracks5-6}) shows the standard deviation of the MC 100 runs in total velocity errors averaged on tracks 5 and 6 while the DLV was not available between 180-200 seconds. ANEKF developed significantly more severe errors than ANUKF.

\section{Conclusion}\label{conc_sec}
Varying dynamics and environmental changes require adjustment of the nonlinear modeling of the process noise covariance matrix. To cope with this task, we proposed the adaptive neural UKF. Our end-to-end regression network, ProcessNet, can adaptively estimate the uncertainty reflected in the process noise covariance based only on inertial sensor readings. Two ProcessNet networks with the same structure that differ in their input/output are run in parallel: one for the accelerometer, the other for the gyroscope readings. \\
To demonstrate the performance of our approach, we focused on the INS/DVL fusion problem for a maneuvering AUV. We evaluated our approach compared to the model-based UKF and neural adaptive EKF using a real-world recorded AUV dataset. On the test dataset, we examined the improvement in estimating the AUV velocity vector concerning the VRMSE and the AUV misalignment concerning the MRMSE. In the VRMSE metric, ANUKF improved by $21.3\%$ over UKF and $8.2\%$ over ANEKF. In the MRMSE metric, ANUKF improved by $5.4\%$ over UKF and $32.1\%$ over ANEKF.
We further examined the robustness of ANUKF in situations of DVL outages that often occur in real-world scenarios. We showed that ANUKF VRMSE and MRMSE are considerably better than ANEKF, providing a $46.7\%$ improvement in the VRMSE metric and $31.1\%$ in the MRMSE metric. 
In conclusion, ANUKF offers an accurate navigation solution in normal operating conditions of INS/DVL fusion and demonstrates robustness in scenarios of DVL outages. Thus, it allows planning and operating in challenging AUV tasks, including varying dynamics and environmental changes. \\
Considering the high cost of navigation-grade inertial sensors, our novel approach may enable the use of tactical-grade IMU on AUV with the possibility of INS/DVL fusion for other marine robotic systems.\\


\bibliographystyle{IEEEtran}
\bibliography{pIBio}
\vskip 0pt plus -1fil
\end{document}